\def\year{2020}\relax
\documentclass[letterpaper]{article} % DO NOT CHANGE THIS
\usepackage{aaai20}  % DO NOT CHANGE THIS
\usepackage{times}  % DO NOT CHANGE THIS
\usepackage{helvet} % DO NOT CHANGE THIS
\usepackage{courier}  % DO NOT CHANGE THIS
\usepackage[hyphens]{url}  % DO NOT CHANGE THIS
\usepackage{graphicx} % DO NOT CHANGE THIS
\usepackage{graphicx}  % DO NOT CHANGE THIS
\usepackage{mathrsfs} %added by jxz
\usepackage{amsfonts,amssymb}  %added by jxz
\usepackage{tabularx} %added by jxz
\usepackage{url}

\usepackage{array}
\newcommand{\PreserveBackslash}[1]{\let\temp=\\#1\let\\=\temp}
\newcolumntype{C}[1]{>{\PreserveBackslash\centering}p{#1}}
\newcolumntype{R}[1]{>{\PreserveBackslash\raggedleft}p{#1}}
\newcolumntype{L}[1]{>{\PreserveBackslash\raggedright}p{#1}}

\usepackage[colorlinks,
            linkcolor=black,
            anchorcolor=black,
            citecolor=black,
            urlcolor=black]{hyperref}

\title{DualVD: An Adaptive Dual Encoding Model for Deep Visual Understanding in Visual Dialogue}
\author{Xiaoze Jiang\textsuperscript{\rm 1,2} Jing Yu \textsuperscript{\rm 1}\thanks{Corresponding authors: Jing Yu and Zengchang Qin.}  Zengchang Qin\textsuperscript{\rm 2*} Yingying Zhuang\textsuperscript{\rm 1,2} Xingxing Zhang\textsuperscript{\rm 3} Yue Hu\textsuperscript{\rm 1} Qi Wu\textsuperscript{\rm 4} \\
  \textsuperscript{\rm 1}Institute of Information Engineering, Chinese Academy of Sciences, Beijing, China\\ \textsuperscript{\rm 2} Intelligent Computing and Machine Learning Lab, School of ASEE, Beihang University, Beijing, China\\  \textsuperscript{\rm 3} Microsoft Research Asia, Beijing, China\\
   \textsuperscript{\rm 4}  University of Adelaide, Australia \\
   \{yujing02, huyue\}@iie.ac.cn,   \{xzjiang,zcqin\}@buaa.edu.cn, xizhang@microsoft.com, qi.wu01@adelaide.edu.au
}

\begin{document}

\maketitle

\begin{abstract}
Different from Visual Question Answering task that requires to answer only one question about an image, Visual Dialogue involves multiple questions which cover a broad range of visual content that could be related to any objects, relationships or semantics. The key challenge in Visual Dialogue task is thus to learn a more comprehensive and semantic-rich image representation which may have adaptive attentions on the image for variant questions. In this research, we propose a novel model to depict an image from both visual and semantic perspectives. Specifically, the visual view helps capture the appearance-level information, including objects and their relationships, while the semantic view enables the agent to understand high-level visual semantics from the whole image to the local regions. Futhermore, on top of such multi-view image features, we propose a feature selection framework which is able to adaptively capture question-relevant information hierarchically in fine-grained level. The proposed method achieved state-of-the-art results on benchmark Visual Dialogue datasets. More importantly, we can tell which modality (visual or semantic) has more contribution in answering the current question by visualizing the gate values. It gives us insights in understanding of human cognition in Visual Dialogue.
\end{abstract}

\section{Introduction}

\begin{figure}[t]
\centering
\includegraphics[width=8cm]{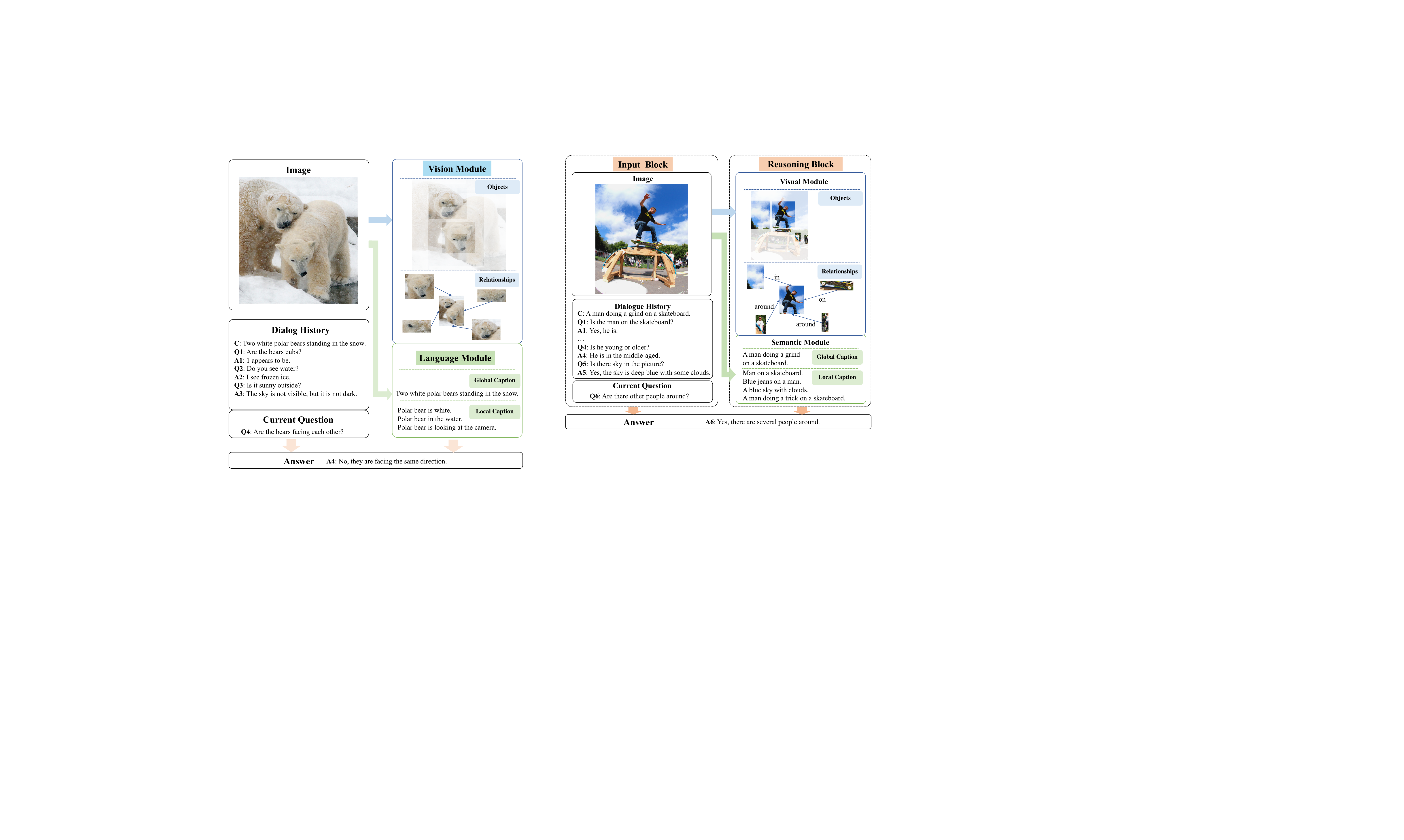}
\caption{An illustration of DualVD. Left: the input of the dialogue system. Right: visual and semantic modules designed to adaptively understand the visual content like humans. The answer is inferred depending on multi-modal evidence.}
\label{basicIdea}
\end{figure}

To understand the real world by analyzing vision and language together is a priority for AI to achieve human-like abilities, which enables the development of diverse applications, such as Visual Question Answering (VQA) \cite{Agrawal2017VQA}, Referring Expressions \cite{Wang2019Neighbourhood}, Image Captioning \cite{johnson2016densecap}, \emph{etc}. To move a step further, this work focuses on the Visual Dialogue \cite{Das2017Visual} problem, which requires the agent to answer a series of questions in natural language regarding an image. %which attracts increasing attention because of its capability to process information across modalities. 
It is more challenging because it demands the agent to adaptively focus on diverse visual content with respect to the current question, while other vision-language problems mostly attend to some specific objects or regions. Considering the dialogue in Figure \ref{basicIdea}: Given `` \emph{Q1: Is the man on the skateboard?}'', the agent should be aware of the foreground visual content, i.e. \emph{the man, the skateboard}, while `` \emph{Q5: Is there sky in the picture?}'' changes the attention of the agent to the background of \emph{sky}. Besides appearance-level questions like \emph{Q1} and \emph{Q5}, `` \emph{Q4: Is he young or older?}'' requires the agent to reason about the visual content for higher-level semantics. How to adaptively capture the desired visual content through dialogue becomes one of the most critical challenges in visual dialogue.  
%Different from other vision-language problems that Visual Question Answering (VQA), where the model answers a single question about an image, visual dialogue systems require answering a follow-up question in light of previous dialogue. The pivotal challenge in visual dialogue is to dynamically capture visual evidence relevant to the on-going question and continue to infer correct answers.

The typical solution for visual dialogue is to firstly fuse visual (\emph{i.e.} image) features %learned from Convolutional Neural Networks (CNN) 
and textual (\emph{i.e.} dialogue history, current question) features %learned from Recurrent Neural Networks (RNN) 
together and then to infer the correct answer. Most approaches focus on enhancing the textural representations by recovering the dialogue relational structure \cite{Zheng2019Reasoning}, imperfect dialogue history \cite{yang2019making}, and dialogue consistency \cite{wu2018areyou}. %the role of answer \cite{guo2019image}, etc. 
However, the role of visual information is at present less studied. %, but not necessarily the less important in visual dialogue. 
Existing models simply use CNN \cite{Simonyan2014Very} or R-CNN \cite{Ren2017Faster} to extract visual features and focus on the question-relevant content. Such visual features have limited expressive ability due to the monolithic representations \cite{Wang2019Neighbourhood}. %\cite{Wang2019Neighbourhood}.
%Although some works devise attention mechanism to focus on the question-relevant visual features, they still have limited expressive ability due to the monolithic visual representations. 
On one hand, questions in a visual dialogue refer to a wide range of visual content, %consider more on the visual appearance and 
 including objects, relationships and high-level semantics, which can not be covered by monolithic features. %and thus hard to answer the question referring to the relationship. 
%Considering the question `` \emph{Q1: Is the man on the skateboard?}'' in Figure \ref{basicIdea}, it requires the model being aware of the visual appearance of \emph{the man, the skateboards} as well as their semantic relationship \emph{on}. 
%Compared with visual representations, such relationships can be more accurately conveyed in language form via complex human cognition. 
On the other hand, the referred visual content may change remarkably from visual appearance to high-level semantics through the dialogue, % and become more involved in the conversation. 
which is difficult for monolithic features to capture. 

%though the visual features involve more subtle semantic clues, they have lower level of abstraction compared to the natural language. For instance,  to answer the question "\emph{Is the cat looking at the dog?}", the model just needs to analyze the visual relationships at concept level (i.e. concept \emph{cat} and \emph{dog}) and that much visual features may introduce unexpected noise. 

Our work is inspired by the Dual-coding theory \cite{paivio1971imagery} of human cognition process. %One theory \cite{grady1998neural,Cermak1979Levels,Craik1975J,paivio1971imagery} underlying superior visual cognition suggests that our brain automatically encodes the images by multiple representations and associates with other knowledge about the world%, thus encouraging a more elaborate encoding than single modal representation. 
\textit{Dual-coding theory}  postulates that our brain encodes information in two ways: \textit{visual imagery} and \textit{textual associations}.  %For example, say a person has stored the concept ``cat'' as both the word ``cat'' and as the image of a cat. 
When asked to act upon a concept, our brain retrieves either images or words, or both simultaneously. The ability to encode a concept by two different ways strengthens the capacity of memory and understanding. % compared to only encode in one way. 
Inspired by the cognitive process, we first propose a novel scheme to comprehensively depict an image from  both  visual  and  semantic  perspectives, where the major objects and their relationships are kept in the visual view while the higher-level abstraction is provided in the semantic view. We propose a model called \textit{Dual Encoding Visual Dialogue} (\textit{DualVD}) to adaptively select question-relevant information from the image in a hierarchical mode: intra-modal selection first captures the visual and semantic information individually from the object-relational visual features and global-local semantic features; then inter-modal selection obtains the joint visual-semantic knowledge by correlating vision and semantics. This hierarchical framework imitates human cognition process to capture targeted visual clues from multiple perceptual views and semantic levels.

%DualVD encodes the visual information by disparate visual and semantic modules, which are coordinated to perform subtle and abstract cognition simultaneously. Visual module extracts object appearance and their visual relationships  %and encodes the subtle visual properties as well as the ``soft-embedded'' relational information. The visual regions and their relationships are organized as a scene graph, which resembles the visual reasoning process regarding to the question.
%while the semantic module explicitly converts the image into high-level global and local textural descriptions. %which  distills the key concepts and their relationships from the language view. which is unified with the question and dialogue history in text domain for correlating the semantics. 
%By dynamically selecting visual evidence from both visual and semantic views, DualVD adaptively derives the visual information for the on-going questions, thus generating more human-like responses and being more evolved in the conversation.  

The main contributions are summarized as follows: %We propose a new thought of representing visual information for the visual dialogue problem. Instead of considering visual features only, 
(1) We exploit the possibility of cognition in visual dialogue by depicting an image from both visual and semantic views, which covers a broad range of visual content referred by most of questions in the visual dialogue task; %The visual representations can also be easily extended to other modalities for deeper visual understanding.
(2) We propose a hierarchical visual information selection model, which is able to progressively select question-adaptive clues from intra-modal and inter-modal information for answering diverse questions. 
% We propose to model the image by a relation-aware scene graph and dynamically enrich the visual features to adapt to the current question. To our best, we are the first to study the role of fine-grained visual representation for better visual dialogue response; 
%(3) Another advantage of DualVD lies in its interpretability. 
It supports explicit visualization in visual-semantic knowledge selection and reveals which modality has more contribution to answer the question; %Extensive experiments verify the significance of incorporating vision and language to guide the answer inference.
(3) The proposed model outperforms state-of-the-art approaches on benchmark visual dialogue datasets, which demonstrates the feasibility and effectiveness of the proposed model. The code is available at \url{https://github.com/JXZe/DualVD}.

\section{Related Work}

\subsubsection{Visual Question Answering (VQA)} focuses on answering arbitrary natural language questions conditioned on an image. %Due to its potential applications in virtual assistants and autonomous agents, it attracts a large quantity of interests in natural language processing and computer vision field. 
The typical solutions in VQA build multi-modal representations upon CNN-RNN architecture \cite{ren2015exploring,wu2017image}. %They adopt deep Convolutional Neural Networks (CNNs) to represent images and Recurrent Neural Networks (RNNs) to represent questions. The extracted visual and textual feature vectors are then jointly embedded %by concatenation, element-wise sum or product 
%to infer the answer. %It would be even complex if extra knowledge is required to be combined for reasoning. \citeauthor{li2017incorporating}  proposed to embed knowledge in memory slots and incorporated external knowledge with image, question and answer features by Dynamic Memory Networks (DMN). 
% de by jxz
%One of the key challenges in VQA is to effectively understand and extract visual features that better adapts to the question. 
Existing approaches incorporate context-aware visual features. 
%and the popular trend for modeling the visual context is progressively from global level to fine-grained level. 
For example, \cite{ren2015exploring} applies CNN features of the whole image as global context, \cite{xu2016ask,anderson2018bottom-up} adopt patches and salient objects learned by attention mechanism as the region context, and \cite{gao2018question,li2019relation} exploits inter-object relationships via graph attention networks or convolutional networks to model the relational context. %Thereinto, the spatial \cite{teney2017graph} or implicit relationships \cite{cadene2019murel} are proposed and proved to be effective in this task. 
However, how to leverage the external visual-semantic knowledge to learn more informative relational representations for better semantic understanding has not been well exploited yet.   
Another emerging line of work represents visual content explicitly by natural language and solves VQA as a reading comprehension problem. In \cite{Li2019Visual}, the image is wholly converted into descriptive captions, which preserves information at semantic-level in textual domain. 
%in the same way as the questions. 
However, this kind of approaches use the generated captions, which could not be correct as we desired, and that they fully abandon the informative and subtle visual features.
Besides the specific tasks, our model has notable progress compared to the above approaches. %First, we encode the visual information by both visual and textural modalities, which is one step further to perform subtle and abstract cognition simultaneously. 
We adopt dual encoding mechanism to provide both appearance-level and semantic-level visual information, so that it incorporates the strong points of the above two kinds of approaches. 
%First, our model provides hierarchical level of representations from different modalities to dynamically capture question-relevant visual clues. Second, we emphasize both relationship learning (from external knowledge) and visual relational reasoning for better relational context representations.

%Not only can they tackle observation based questions, but also can handle knowledge-based VQA problems. Replacing image with caption, the model only depends on the quality of caption generation.Inaccurate caption will introduce noise to the system. Besides, lacking of fine-grained semantic information in images, text only contains the concept information. Our work realize the complementary role of the two kinds of information for visual content understanding and establish Multi-Modal representations. 

\subsubsection{Visual Dialogue} %is a recently proposed vision-grounded NLP task as a combination of visual grounding, image captioning, and visual question answering. Specifically, 
aims to answer a current question conditioned on an image and dialogue history. 
%\citeauthor{Das2017Visual}(2017) released the VisDial dataset containing free-form natural language questions and answers related to a given image. They further formulated the multi-round VQA task. Most researches on the task follow the feature fusion framework, then attention mechanism is combined to strengthen the interaction between visual and contextual features. 
Most existing works are based on late fusion framework and focused on modeling the dialogue history. Sequential co-attention mechanism \cite{wu2018areyou} enables the model to identify question-relevant image regions and dialogue history to keep the dialogue consistency. \cite{yang2019making} introduces false response in dialogue history for an adverse critic on the historic error.  %\citeauthor{guo2019image} proposed a novel image-question-answer synergistic network to value the role of the answer for precise visual dialogue. 
%To investigate semantic dependencies between entities underlying dialogue,  
\cite{Zheng2019Reasoning} introduces an Expectation Maximization algorithm to infer the dialogue structure and the answers via graph neural networks. 
%Specifically, given dialogue entities are viewed as the observed nodes, the answer to a given question is represented by a node with missing value and the edge weights are represented as the semantic dependencies between nodes. The problem is formulated as inferring graph structures (relations in dialogue) and the missing values. In this way, not only the reasoning process utilizes the rich relational information in this task, but also iteratively and deeply mines the information from different dialogue entities. 
By contrast to extensive study on modeling dialogue history, the image content has been less studied. Although some works devise attention mechanism to focus on the essential visual features most relevant to the question and dialogue history, such monolithic visual representations still have limited expressive abilities. In this work, we exploit the role of visual information in visual dialogue. Different from existing works merely modeling the  appearance, our model is able to adaptively capture visual and semantic information in a hierarchical mode inspired by the Dual-coding theory of human cognition process to provide adequate visual clues for diverse questions in visual dialogue.

\begin{figure*}[t]
\centering
\includegraphics[width=17.5cm]{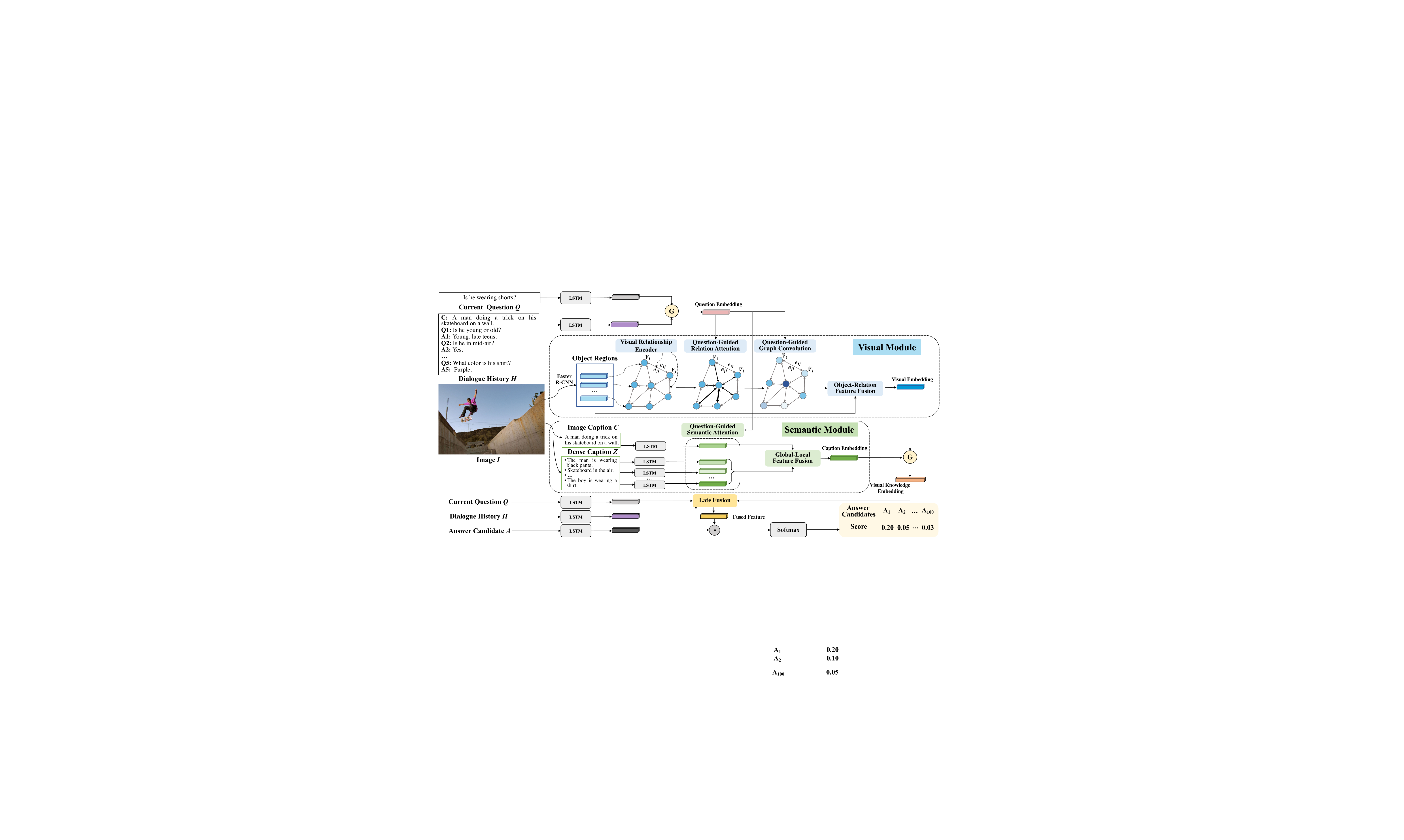}
\caption{Overview structure of the DualVD model for visual dialogue. The model mainly contains two parts: Visual Module and Semantic Module, where ``G'' represents gate operation given inputs.}
%{Overview of the proposed DualVD model. We use Vision Module and Language Module to encode the Image \textsl{I}. Specifically, in Vision Module, we build a scene graph between object resigns (as nodes) and visual relationships (as edges), and assign question (updated by history) to update the appearance-level visual information. In Language Module, we fuse the Global Caption (provided by dataset) and Local Caption (extracted by DenseCap) to get the semantic-level image information. Then the representation of image is updated by Visual Embedding and Caption Embedding. We use LSTM to encode History \textsl{H} and Question \textsl{Q}, and feed the Fused Feature (joint \textsl{I}, \textsl{H} and \textsl{Q}) to Decoder.}
\label{model_pic}
\end{figure*}

\section{Methodology}
The visual dialogue task can be described as follows: given an image $I$ and its caption $C$, a dialogue history till round $t$-$1$, $H_t = \{ C, (Q_1,A_1),...,(Q_{t-1},A_{t-1})\}$, and the current question $Q_t$, 
the task is to rank a list of 100 candidate answers $\mathbb{A}  = \{ A_1, A_2,...,A_{100}\}$ and return the best answer $A_t$ to $Q_t$. %Our model is built on the basic late fusion (LF) framework \cite{Das2017Visual}. 
In this section, we first introduce the idea of  depicting an image from both visual and semantic perspectives. It covers a broad range of visual content like objects, relationships, global semantics and local semantics. Then we introduce a hierarchical feature selection approach to adaptively capture question-relevant visual-semantic information. %the way which we adopted to select the image information from different level adaptively. 
Our model is based on the late fusion (LF) framework \cite{Das2017Visual}, which will be described at the end of this section. %Thus, guided by the dialogue history and the current question, the proposed model DualVD can dynamically selects information from both the fine-grained level and the visual-semantic level to provide effective clues for answer inference.

%the modules in late fusion (LF) framework \cite{Das2017Visual} that will be used in our model. Then we propose DualVD to tackle visual dialogue by encoding the image by two ways: visual imagery (vision module) and textual semantics (language module). Specifically, the \textbf{Vision Module} is responsible for modeling visual objects and their relationships, which are the appearance-level visual information and the primary attentions of human on an image; and the \textbf{language module} converts the image into a global-local textural description, so as to preserve  semantic-level visual information and provide abstract and unified representation with the question in textural domain. Guided by the dialogue history and the current question, the model dynamically selects information from both the vision module and language module to provide effective clues for answer inference.

\subsection{Visual-Semantic Dual Encoding}
\label{DualEncoding}

In visual dialogue, two types of information play the primary role to depict an image and answer the diverse questions: visual information and semantic information (Figure \ref{model_pic}). For visual information, the major objects and relationships should be kept. %in the form of visual features. 
In semantic information, higher-level abstraction of the image content should be provided, which involves prior knowledge and complex cognition. %Considering the above information together, the agent is capable to answer most questions from various perspectives. 
In this section, we introduce a dual encoding scheme to generate both visual and semantic representations to depict an image. A scene graph is proposed to represent the visual information %, where nodes indicates objects while edges depict their relationships (e.g. interactive, compositional, geometric, etc) . In terms of semantic information, we propose to convert an image into 
while multi-level captions in natural language %, so as to implicitly introduce human inference for deep visual understand. 
are leveraged to represent the semantic information. These representations are served as the input of our DualVD model. %where global caption is used to convey the semantics of the whole image while dense captions provides fine-grained semantics over local regions.The question we met in visual dialogue is more complex than other similar tasks (VQA, etc.) and mainly divided into visual-relevent and semantic-relevent two types. To this end, we encode the image information in two different level, the one is appearance-visual level, the other is semantic-level. Considering that a majority of visual-relevent questions are referring to the visual objects and their semantic relationships, we exploit to effectively represent the object appearance and their relational semantics in an image. And the semantic-relevent questions are referring to the more abstract problems. Some are referring to the global image information, others are local. In the meantime, among all the modalities, natural language is a flexible one to describe the high-level semantic visual information via complex cognitive process from the perceived low-level visual features. A diverse range of prior knowledge and reasoning ability is involved in the transformation from vision to language. We attempt to convert the image into hierarchical descriptions in natural language, so as to implicitly introduced prior knowledge and human inference for deep visual understand.

\subsubsection{Scene Graph Construction} Each image is represented as a scene graph. Let $V=\{v_i\}^{N}$ denotes its nodes, which represents objects detected by a pre-trained object detector and let $E=\{e_{ij}\}^{N\times N}$ denotes its edges, which represents the semantic visual relationships embedded by our visual relationship encoder. 
%(more details can be found in the next paragraph). 
We use a pre-trained Faster-RCNN \cite{Ren2017Faster} to detect $N$ objects in an image and describe the object $v_i$ as a $2,048$-dimensional vector, denoted by $h_i$. The visual relationship encoder \cite{zhang2019large}, which is pre-trained on a visual relationship benchmark, i.e. \textit{GQA} \cite{hudson2019GQA}, encodes relationships between the subject $v_i$ and object $v_j$ as a $512$-dimensional relation embedding, denoted as $r_{ij}$. We assume that certain relationship exists between any pair of objects by considering ``unknown-relationship'' as a special kind of relationship. Therefore, the scene graph we constructed is fully-connected.

The visual relationship encoder embeds the relationships between objects into a semantic space which is aligned with their corresponding descriptions in natural language. Such continuous representations instead of discrete labels can preserve the discriminative capability and contextual awareness. 
%The relationships between visual objects are essential to understand image information. %It should be note that there is existing a lot of work to build the relationship between different visual entities. However, in real-word scenarios with large numbers of objects and relations, some are seen very commonly while others are barely seen, and a lot of work failed to represent the complex visual relationship. 
Inspired by recent work \cite{zhang2019large}, our encoder consists of a visual part and a textual part. The visual part takes three CNN feature maps corresponding to the visual regions of subject, object and their union region as input and outputs the three encoded embeddings $x^s$, $x^o$ and $x^r$. The textual part uses a shared GRU to encode the annotations and yield textual embeddings. % $y^s$, $y^o$ and $y^p$.
The loss function is designed to minimize the cosine similarity between the embeddings of positive visual-textual pairs and alienate negative pairs.  The union embedding $x^r$ is served as the visual relationship representation $r_{ij}$ between $v_i$ and $v_j$.

\subsubsection{Multi-level Image Captions} %We propose to represent the semantics of an image via captions from both global and local levels. 
The advantages of captions compared to visual features lie in that captions are represented by natural language with high-level semantics, which can provide straightforward clues for the questions without ``heterogeneous gap''. %by both global-level abstraction covering the whole image and the local-level dense captions for salient regions. 
Global image caption $C$ (provided by the dataset) is beneficial to response to questions exploring the scene. Meanwhile, dense captions \cite{johnson2016densecap}, denoted as $Z=\{z_1, z_2,...,z_k\}$ ($k$ is the number of dense captions), provide a set of local-level semantics, including the object properties (position, color, shape, $etc.$), the prior knowledge related to the objects (weather, species, emotion, $etc.$), and the relationships between objects (interactions, spatial positions, comparison, $etc.$). %, so as to support answering questions digging deeper into the local parts or properties.
The words in both $C$ and $Z$ are represented by concatenated GloVe \cite{Pennington2014Glove} and ELMo \cite{Peters2018Deep} word embeddings. Then $C$ and $Z$ are separately encoded with two different LSTMs, denoted as $\widetilde{C}$ and $\widetilde{Z}=\{\widetilde{z}_1, \widetilde{z}_2,...,\widetilde{z}_k\}$, respectively.

\subsection{Adaptive Visual-Semantic Knowledge Selection}
\label{selection}

On top of the visual and semantic image representations, we propose a novel feature selection framework to adaptively select question-relevant information from the image. Under the guidance of the current question, the feature selection process is devised in a hierarchical mode: intra-modal selection first captures the visual and semantic information respectively from the \textit{visual module} and \textit{semantic module}; then inter-modal selection obtains the desired visual knowledge from both the visual module and semantic module via \textit{selective visual-semantic fusion}. The advantages of such hierarchical framework is that it can explicitly reveal the progressive feature selection mode and preserve fine-grained information as much as possible.

%we propose a framework to select information from different aspect(visual and semantic) and different level(fine-grained level and visual-semantic high level). The framework mainly consists of visual module and semantic module. Visual module selects appearance-level visual information by updating the object with relationship and question. Semantic module via attention mechanism and gate fusing to select the semantic-level image information from global and local level. Then the two different style image information selected by a gate to decide which information is more important to answer the current question.

\subsubsection{Visual Module}

This module is presented on the top of Figure \ref{model_pic}. %Each image is represented as a scene graph with each node denoting an object and each directed edge denoting the relationship between objects. 
Based on the constructed scene graph introduced in \textit{Scene Graph Construction}, we aim to select question-relevant relation information and object information. For relation information, we propose a relation-based graph attention network to enrich the object representations with question-aware relationships. It mainly consists of two units: \textit{Question-Guided Relation Attention} highlights the critical relationships %identifies the critical relationships to answer the question; 
and \textit{Question-Guided Graph Convolution} enriches the object features by its relation-critical neighbors. 
For object information, we highlight the most informative objects to answer the question. Finally, the clues of objects and relationships are further fused in \textit{Object-Relation Information Fusion} to obtain the question-relevant visual content. 

\textit{Question-Guided Relation Attention}: The question-guided relation attention examines all the relationships to highlight the ones most relevant to the question. First, we select question-relevant information from the dialogue history to merge into the question representation via a gate operation, which is defined as:

\begin{equation}
gate_t^q =  \sigma (\textbf{W}_q [\widetilde{H}_t, \widetilde{Q}_t] + b_q)
\label{gate_q}
\end{equation}
\begin{equation}
\widetilde{Q}_{t}^g =  \textbf{W}_1  (gate_t^q \circ  [\widetilde{H}_t, \widetilde{Q}_t] )+ b_1
\label{q_h}
\end{equation}
where ``$[\cdot ,\cdot ]$'' denotes concatenation, ``$\circ$'' denotes the element-wise product. Each word is represented by concatenating the hidden states extracted from pre-trained GloVe and ELMo models. Then dialogue history $H_t$ and the current question $Q_t$ are separately encoded with two different LSTMs, denoted as $\widetilde{H}_t$ and $\widetilde{Q}_t$, respectively. $gate_t^q $ is a vector of  gate values over $\widetilde{H}_t$ and $\widetilde{Q}_t$, $\textbf{W}_1$ (as well as $\textbf{W}_2, ..., \textbf{W}_7$ mentioned below) is the linear transformation layer and $\widetilde{Q}_{t}^g$ is the encoded history-aware question features.

The attention weights $\alpha _{ij} $ of all the visual relationships are calculated under the guidance of the question $\widetilde{Q}_{t}^g$:
\begin{equation}
\alpha _{ij} = softmax(\textbf{W}_\rho  (\textbf{W}_2   \widetilde{Q}_{t}^g  \circ \textbf{W}_3  r_{ij}) + b_{r})
\end{equation}

Each relation embedding is updated based on the attention importance. Formally defined  as:
\begin{equation}
\widetilde{r}_{ij} = \alpha _{ij}  r_{ij}
\end{equation}
where $\widetilde{r}_{ij}$ is the question-guided relation embedding.

%write jxz
\textit{Question-Guided Graph Convolution}: This module further updates each object's representation under the guidance of questions by aggregating information from its neighborhood and the corresponding relationships. Given the feature $h_j$ of object $v_j$ and its relation embedding $\widetilde{r}_{ij}$, the attention value of $v_j$ w.r.t. $v_i$ is calculated as:

%For each object node $v_i$, we first contact its neighbor nodes $v_j$ and their relationship $e_{ij}$, and then use the question $Q_{t}^g$ to get the convolution weight $\beta_{ij}$:

\begin{equation}
\beta _{ij} = softmax(\textbf{W}_{g}  ( \widetilde{Q}_{t}^g \circ ( \textbf{W}_4 [h_j,\widetilde{r}_{ij}]) )+ b_{g})
\label{relationAtt}
\end{equation}

The obtained attention values for all the neighbors of $v_i$ are used to compute a linear combination of their features, which serves as the updated representation $\widetilde{h}_i$ for $v_i$:

\begin{equation}
\widetilde{h}_i = \sum_{j=1}^{N}\beta_{ij} h_j
\end{equation}

Since the scene graph is a fully connected graph, the number of neighbors $N$ for each object is equal to the number of objects detected in each image. 

%write jxz
\textit{Object-Relation Information Fusion}: In visual dialogue, the object appearance and the visual relationships will contribute to infer the answer, but with different contributions. In this module, we adaptively fuse question-relevant object features  from both original object feature $h_i$ and relation-aware object feature $\widetilde{h}_i$ again by a gate, 
%For some questions, agents only know appearance visual information can answer correctly, such as `` \emph{what color is the T-shirt?}''. Object relation feature fusion dynamically control the fusion ratio of original nodes $v_i$ and updated nodes $v_i^e$ via a gate, 
which is defined by:
\begin{equation}
gate_i^v =  \sigma (\textbf{W}_{v}  [h_i, \widetilde{h}_i] + b_{v})
\label{gate_v}
\end{equation}
\begin{equation}
\widetilde{h}_i^g =  \textbf{W}_5  ( gate_i^v \circ [h_i, \widetilde{h}_i]) +b_5
\label{q_h}
\end{equation}
where $\widetilde{h}_i^g$ is the updated representation of object $v_i$. The whole image representation $\widetilde{I}$ is obtained as the weighted sum of the object representations.  %We also noticed the importance of the origin question information, for the question is the most relevant textual information when make dialogue and the fusion of question and history may weakness the role of question. 
In order to strengthen the influence of the current question $Q_t$ and the original object features on the retrieved visual clues, we calculate the attention value $\gamma_i^v$ for $h_i$ under the guidance of $Q_t$: %and then to update the whole representation of the image $\widetilde{I}$:
\begin{equation}
\gamma_i^v =  softmax (\textbf{W}_{s}  (Q_t \circ ( \textbf{W}_6  h_i))+ b_{s})
\label{objectAtt}
\end{equation}
Then the the whole representation of the image $\widetilde{I}$ can be updated by:
\begin{equation}
\widetilde{I} = \sum_{i=1}^{N}\gamma_i^v \widetilde{h}_i^g
\end{equation}

\subsubsection{Semantic Module}

This module aims to select and merge question-relevant semantic information from global and local captions with a \textit{Question-Guided Semantic Attention} module and a \textit{Global-Local Information Fusion} module. The semantic module is located in the middle of Figure \ref{model_pic}.

\textit{Question-Guided Semantic Attention}: The semantic attention mechanism highlights relevant captions at both global-level and local-level. This type of attention is guided by the current question which is enhanced with corresponding information from the dialogue history (as introduced above). According to the attention distribution, we enrich the caption representations in order to better adapt to the question. The attention value for each caption in $m_i \in \{\widetilde{C},\widetilde{z}_1, \widetilde{z}_2,...,\widetilde{z}_k\}$ is calculated as follows:
\begin{equation}
\delta_i^q =  softmax( (\textbf{W}_{z1}  \widetilde{Q}_t^g + b_{z1})^T     (\textbf{W}_{z2}  m_i + b_{z2}) )%softmax (sum((\textbf{W}_{z1} \cdot Q_t^g + b_{z1}) \cdot (\textbf{W}_{z2} \cdot j_i + b_{z2}) ))
\label{languageAtt}
\end{equation}

The caption representation for $\widetilde{C}$ and $\widetilde{Z}$  will be updated to $\widetilde{C}^q$ and $ \widetilde{Z}^q$:
\begin{equation}
\widetilde{C}^q =  \delta_1^q  C
\end{equation}
\begin{equation}
\widetilde{Z}^q =   \sum_{i=2}^{k+1}\delta_i^q  \widetilde{z}_{i-1}
\end{equation}

\textit{Global-Local Information Fusion}: Some questions are global-related while others are local-related. This step adaptively selects the information from the global caption $\widetilde{C}^q$ and local caption $ \widetilde{Z}^q$ via a gate as described above:
\begin{equation}
gate^c =  \sigma (\textbf{W}_{c} [\widetilde{C}^q ,\widetilde{Z}^q] + b_{c})
\end{equation}
\begin{equation}
\widetilde{T} =   \textbf{W}_7 (gate^c \circ  [\widetilde{C}^q ,\widetilde{Z}^q])+b_7
\end{equation}
where $\widetilde{T}$ is the textural representations for the abstract visual semantics.
%This step inspects the information in global caption $\hat{C}$ and local captions $\hat{Z}$ and retrieves the information that is required for inferring current question. For a pair of $\hat{C}$ and each of the local caption $\hat{z}_i$, the content's relevance is measured by a gate mechanism that assign probability to each feature in $\hat{C}$ and $\hat{z}_i$ taking into account of the question $Q$. The attention distribution is calculated 

\subsubsection{Selective Visual-Semantic Fusion}

When asked to answer a question, the agent will retrieve either the visual information or the semantic information individually, or both simultaneously.
%According to the cognitive process of human, people not only use visual information but also use textual information to understand the world. The Visual Model and Semantic Model are two different representations of visual information, and 
In this module, we design a gate operation to decide the contributions of the two modalities on the answer prediction. The gate operation and the final visual knowledge representation $S$ are calculated as:
\begin{equation}
gate^s =  \sigma (\textbf{W}_{s}  [\widetilde{I} ,\widetilde{T}] + b_{s})
\label{fussAtt}
\end{equation}
\begin{equation}
S =  gate^s \circ  [\widetilde{I} ,\widetilde{T}]
\end{equation}

\subsection{Late Fusion and Discriminative Decoder}
\label{LateFusion}
%We adopt late fusion framework, which follows the encoder-decoder framework, to fuse the information in visual dialogue. 
The full model consists of late fusion encoder and discriminative (softmax) decoder. %The details can refer to \cite{Das2017Visual}.  
%\textbf{Input Embedding and Late Fusion Encoder}: 
The encoder first embeds each part in a dialogue tuple $D=\{I, H_t, Q_t\}$. %The dialogue history $H_t$ is treated as a long string by concatenating the image caption $C$ with previous $t-1$ round dialogues. 
Then we concatenate  $\widetilde{H}_t$ and $\widetilde{Q}_t$ with the visual knowledge representation $S$ into a joint input embedding for answer prediction. 
%For the image $I$, previous model extracts the global embedding with CNNs. %after the L2-normalized activations from the penultimate layer of VGG-16 as the visual representation. 
%Considering that diverse range of question-specific visual information can't be comprehensively described by the global monolithic features, our work describe the image by both relation-aware visual imagery and concept-aware textural association, %so as to capture comprehensive semantic information.
%which will be illustrated in the following sections. %Section \ref{VisionModule} and Section \ref{LanguageModule}. 
%The representations in the input tuple $D$ are concatenated and mapped into a joint representation for answer prediction. 
%characterizes the text from the word level and character level and uses LSTM to extract semantic information for finding a better vector representation of the text in the semantic space.
%\textbf{Deep Visual Module} processes visual information in depth, including image feature extraction, dense caption generation, visual relation graph building and the updating of visual entities based on visual relation graph and question.
%\textbf{Late Fusion Encoder} fuses the information from different modal, including the different representations for visual information and the representations from question and history, and then puts the joint representation to a discriminative decoder to predict the answers.
%\textbf{Discriminative (softmax) Decoder}: 
The decoder ranks all the answers from a set of 100 candidates $\mathbb{A}$. It first encodes each candidate via a common LSTM. Then a dot product followed by softmax operation is calculated between the joint input embedding and candidates to get the posterior probability over each candidate. We obtain the correct answer by ranking the candidates based on their posterior probabilities. Our model can also be applied to more complex decoders and fusion strategies, such as memory network, co-attention, adversarial network, \emph{etc}. In this paper, we utilize the simple late fusion and  discriminative decoder to highlight the advantages of our visual encoder.

\section{Experiments}
\label{sec:experiments}

%In this section, we first present quantitative comparisons between our model and state-of-the-art works. Further, an ablation study shows the influence of the key modules in our model. Finally,  the effectiveness of visual module and semantic module for answering different types of questions are further analyzed by visualization and empirical study. 

\textbf{Datasets}: We conduct extensive experiments on datasets \cite{Das2017Visual}: VisDial v0.9 and VisDial v1.0. For both datasets, the examples are split into  ``train'', ``val'' and ``test'' and 
%the dialogues are collected from the chatting history by two users in Amazon Mechanical Turk (AMT) about each image in MSCOCO. 
each dialogue contains 10 rounds of question-answer pairs. 
%The dialogues are split into  ``train'', ``val'' and ``test''.  In ``train'' and ``val'', each image is accompanied with a 10-round dialogue  while in ``test'', each image is followed by a random rounds of question-answer pairs and an on-going question for answer prediction.  
VisDial v1.0 is an upgraded version of VisDial v0.9. %The difference between the two datasets lies in the data source and splits. 
For VisDial v0.9, all the splits are built on MSCOCO images. 
%including 80k dialogues in ``train'' on COCO-train images, 3k dialogues in ``val'' on COCO-train images and 40k dialogues in ``test'' on COCO-val images. 
For VisDial v1.0, all the splits of VisDial v0.9 serve as ``train'' (120k), while ``val'' (2k) and ``test'' (8k)  consist of dialogues on extra 10k COCO-like images from Flickr. 

\textbf{Evaluation Metrics}: We follow the metrics in \cite{Das2017Visual} to evaluate the response performance. In the test stage, the model is asked to rank 100 candidate answer options and evaluated by Mean Reciprocal Rank (MRR), Recall\textsl{@}$k (k = 1, 5, 10)$ and Mean Rank of human response (Mean) on both datasets. 
%As some of the candidate answer option may be semantically identical (e.g. 'hi' and 'hello') or relevant to some extent, 
For VisDial v1.0, Normalized Discounted Cumulative Gain (NDCG) is added as an extra metric for more comprehensive analysis. Lower value for Mean and higher value for other metrics are desired.

\textbf{Implementation Details}: For the textual part, the maximum sentence length of the dialogue history,  dense captions and the current question is all set to 20. The hidden state size of all the LSTM blocks is set to 512.  We use Faster-RCNN with the ResNet-101 to detect object regions and extract the 2048-dimensional region features.
%from the Res4b22 feature map. 
Since some captions with low confidence are likely to introduce unexpected noise and too many captions will decrease the computation efficiency, we select the top 6 (the mean value of the caption distribution) dense captions in our model. We train all of our models by Adam optimizer with 16 epochs, where the mini-batch size is 15 and the dropout ratio is  0.5. For the strategy of learning rate, %we combine warm-up and cosine annealing learning strategy. In detail, 
we first apply warm up strategy for 2 epoches with initial learning rate  $1\times10^{-3}$ and warm-up factor 0.2. Then we adopt cosine annealing learning strategy with initial learning rate $\eta_{max}$=$1\times10^{-3}$ and termination learning rate $\eta_{min}$=$3.4\times10^{-4}$ for the rest epoches. %All the experiments are conducted on an NVIDIA Titan X GPU with 12 GB memory and implemented in PyTorch.
\begin{table}[t]
\caption{Comparison on validation split of VisDial v0.9.}
\label{v9}
%\begin{center} 
\resizebox{.95\columnwidth}{!}{
\begin{tabular}{L{2.42cm}C{0.68cm}C{0.68cm}C{0.68cm}C{0.68cm}C{0.68cm}}%{llllllX}   
%\begin{tabular}{L{1.5cm}L{0.65cm}L{0.65cm}L{0.65cm}L{0.65cm}L{0.65cm}L{0.7cm}}
\hline                       
Model & MRR & R\textsl{@}1 & R\textsl{@}5 & R\textsl{@}10 & Mean \\
\hline  
LF &58.07&43.82&74.68&84.07&5.78  \\
HRE &58.46 &44.67&74.50&84.22&5.72\\  
%HREA &58.68&44.82&74.81&84.36&5.66 \\  
MN & 59.65 & 45.55 & 76.22&85.37& 5.46 \\ 
SAN-QI &57.64&43.44&74.26&83.72&5.88\\
HieCoAtt-QI &57.88&43.51&74.49&83.96&5.84\\
AMEM &61.60&47.74&78.04&86.84&4.99\\
HCIAE &62.22&48.48&78.75&87.59&4.81\\
SF &62.42&48.55&78.96&87.75&4.70\\
CoAtt &63.98&50.29&80.71&88.81&4.47\\
CorefMN &\textbf{64.10} & \textbf{50.92} &80.18&88.81&4.45\\
VGNN &62.85&48.95&79.65&88.36&4.57\\
%FGA & \textbf{65.25} & \textbf{51.43} & \textbf{82.08} & 89.56 & 4.35 \\
\hline
\textbf{DualVD}  & 62.94 & 48.64 & \textbf{80.89} & \textbf{89.94} & \textbf{4.17} \\
%\textbf{DualVD-MN}  & 63.03 & 48.75 & 81.09 & \textbf{90.27} & \textbf{4.12} \\
\hline  
\end{tabular}  
}
%\end{center} 
\end{table}

\subsection{Overall Results}

%In this section, we show the comparison results  between our model and state-of-the-art approaches on VisDial v0.9 and VisDial v1.0. 
In Table \ref{v9} and Table \ref{v1}, we compare DualVD with state-of-the-art discriminative models, namely  LF \cite{Das2017Visual}, HRE \cite{Das2017Visual}, 
%HREA \cite{Das2017Visual}, 
MN \cite{Das2017Visual}, SAN-QI \cite{YangZi2016Stacked}, HieCoAtt-QI \cite{Lu2016Hierarchical}, AMEM \cite{seo2017visual},  HCIAE \cite{lu2017best}, SF \cite{Jain2018Two}, CoAtt \cite{wu2018areyou}, CorefMN \cite{Kottur2018Visual}, VGNN \cite{Zheng2019Reasoning}, %FGA\cite{schwartz2019factor}, 
LF-Att \cite{Das2017Visual}, MN-Att \cite{Das2017Visual}, 
RvA\cite{Niu2018Recursive} and DL-61\cite{guo2019image}. Our model consistently outperforms all the approaches on most metrics, which highlights the importance of visual understanding from visual and semantic modules in visual dialogue. CoAtt and HeiCoAtt-QI are relevant to our model in the sense that they leverage attention mechanism to identify question-relevant visual features. However, they ignore the semantic-rich relationships and language priors. %We consider both object-relation visual information and global-local semantic information relevant to the question and achieve richer image understanding. 
It should be noted that our model and the compared approaches all belong to single-step models. With the success of multi-step reasoning, %the recent proposed 
ReDAN \cite{Gan2019Multi} achieves 1\% boost over our model on most metrics. We believe that stacking our visual encoder to achieve multi-step visual understanding is a promising future work. DL-61 \cite{guo2019image} is a two-stage network for candidate selection and re-ranking while FGA \cite{schwartz2019factor} conducts attention across all the data parts, which gain relatively high performance on some metrics compared with our model. We believe that our model for the visual part and existing works for the dialogue or answer parts have complementary advantages. 

%To prove the complementary advantage of our visual representation on the ‘dialog history’ models, we conducted an experiment to apply the proposed visual representation on top of the Memory Network baseline \cite{Das2017Visual}, which pays more attention on ‘dialog history reasoning’. The results ( \textbf{DualVD-MN} in Table \ref{v9} and Table \ref{v1}) gain further improvement over our model on most metrics. It proves the complementary contribution of our model on the existing ‘dialogue history’ models. 

%Our model consistently outperforms all the approaches on almost all metrics, which highlights the importance of visual understanding from visual and semantic modules in visual dialogue. The representative CoAtt and HeiCoAtt-QI are relevant to our model in the sense that they leverage attention mechanism to identify question-relevant visual features. However, they ignore the semantic-rich relationships and language priors. We consider both object-relation visual information and global-local semantic information relevant to the question and achieve richer image understanding. To notice that our model and the compared approaches all belong to single-step models. With the success of multi-step reasoning, the recent proposed ReDAN \cite{Gan2019Multi} achieves 1\% boost over our model on most metrics. We believe that stacking our visual encoder to achieve multi-step visual understanding is a promising future work.

\begin{table}[t] 
\caption{Comparison on test-standard split of VisDial v1.0.}
\label{v1}
%\begin{center} 
\resizebox{.95\columnwidth}{!}{
\begin{tabular}{L{1.66cm}C{0.6cm}C{0.6cm}C{0.6cm}C{0.69cm}C{0.6cm}C{0.77cm}}
%{L{1.66cm}C{0.6cm}C{0.6cm}C{0.6cm}C{0.69cm}C{0.6cm}C{0.77cm}}
%L{1.95cm}C{0.58cm}C{0.58cm}C{0.58cm}C{0.60cm}C{0.57cm}C{0.76cm}
%{llllllX}   
%\begin{tabular}{L{1.5cm}L{0.65cm}L{0.65cm}L{0.65cm}L{0.65cm}L{0.65cm}L{0.7cm}}
\hline                       
Model & MRR & R\textsl{@}1 & R\textsl{@}5 & R\textsl{@}10 & Mean & NDCG  \\
\hline  
LF &55.42 & 40.95 & 72.45 & 82.83 & 5.95 & 45.31 \\
HRE & 54.16 & 39.93 & 70.47 & 81.50 & 6.41& 45.46 \\  
MN & 55.49 & 40.98 & 72.30 & 83.30 & 5.92 &47.50\\  
LF-Att &57.07&42.08&74.82&85.05&5.41&40.76\\
MN-Att & 56.90 & 42.43 & 74.00 & 84.35 & 5.59 & 49.58\\
CorefMN & 61.50 & 47.55 & 78.10 & 88.80 & 4.40 &54.70\\
VGNN & 61.37 &47.33 &77.98&87.83&4.57&52.82\\
RvA &63.03&49.03&80.40&89.83&4.18&55.59\\
DL-61 &62.20&47.90&\textbf{80.43}&\textbf{89.95}&4.17&\textbf{57.32}\\
%FGA &\textbf{63.70}& \textbf{49.58}& \textbf{80.98}&88.55&4.51&52.10\\
%\hline  
%RvA & 63.03&49.03&80.40&89.83&4.18&55.59\\
%ReDAN &64.75& 51.10& 81.73& 90.90& 3.89&57.63\\ 
\hline
\textbf{DualVD} & \textbf{63.23} & \textbf{49.25} & 80.23 & 89.70 & \textbf{4.11} & 56.32\\
%\textbf{DualVD-MN} & 63.64 & \textbf{49.90} & \textbf{81.08} & \textbf{89.98} & 4.14 &56.12\\
\hline  
\end{tabular}  
%\end{center} 
}
\end{table}

\subsection{Ablation Study}

Ablation study on VisDial v1.0 validation set exploits the influence of the essential components of DualVD. We use the same discriminative decoder for all the following variations:

%Specifically, we compare the following variations:
\textbf{Object Representation (ObjRep)}: this model %is the baseline model in Visual dialogue challenge 2019 \cite{desai2019visdialch} and 
uses the averaged object features to represent an image. Object representations are enhanced by question-driven attention.
%$\footnote{ObjectRep(baseline model in Visual dialogue challenge 2019) source code is available from https://github.com/batra-mlp-lab/visdial-challenge-starter-pytorch}$. 
%In baseline model, we extract image features via a pre-trained Faster R-CNN model and do attention between Question and Image. For text information, we first randomly initialize the word vector using the traditional word2vec method and then encode Question and History via a LSTM respectively. In the end, we contact the representation of Image, Question and History to a vector and then feed them to decoder.

\textbf{Relation Representation (RelRep)}: %on the top of object representation, 
this model applies averaged relation-aware object representations via \emph{question-guided relation attention} and \emph{question-guided graph convolution} as the image representation. %relation representation leverages visual relationship encoder to encode edge semantics and apply relation attention and graph convolution to aggregate neighborhood features for object representation, as introduced in Section Vision Module without \textit{object-relation feature fusion} stage.

\textbf{Visual Module without Relationships (VisNoRel)}: this is our full visual module except that the relation embeddings are replaced by unlabeled edges and the convolution is conducted via the intra-modal attention \cite{gao2019dynamic}.

\textbf{Visual Module (VisMod)}: this is our full visual module, which fuses objects and relation features. 

\textbf{Global Caption (GlCap)}: this model uses LSTM to encode the global caption to represent the image. 

\textbf{Local Caption (LoCap)}: this model uses LSTM to encode the local captions to represent the image. 

\textbf{Semantic Module (SemMod)}: this is our full semantic module, which fuses global and local features. 

\textbf{DualVD (full model)}: this is our full model, which incorporates both the visual module and semantic module.

%\textbf{Baseline + ELMo \& GLoVe (BE)}: On the basis of BL, we replace the word2vec with ELMo \& GLoVe as described in DualVD.

%\textbf{BE + Dense Caption (BED)}: On the basis of BE, we add Dense Caption for each image following the way we used in DualVD.

%\textbf{BEG + Relation Graph (BEG)}: On the basis of BE, we add Relation Graph module as described in DualVD.

%\textbf{BEG - Relation Attention (BEG-Ra)}: On the basis of BEG, we do not use Question to update the representation of relation of each visual entities. Instead, we use origin relation 

\begin{figure*}[t]
\centering
\includegraphics[width=17.6cm]{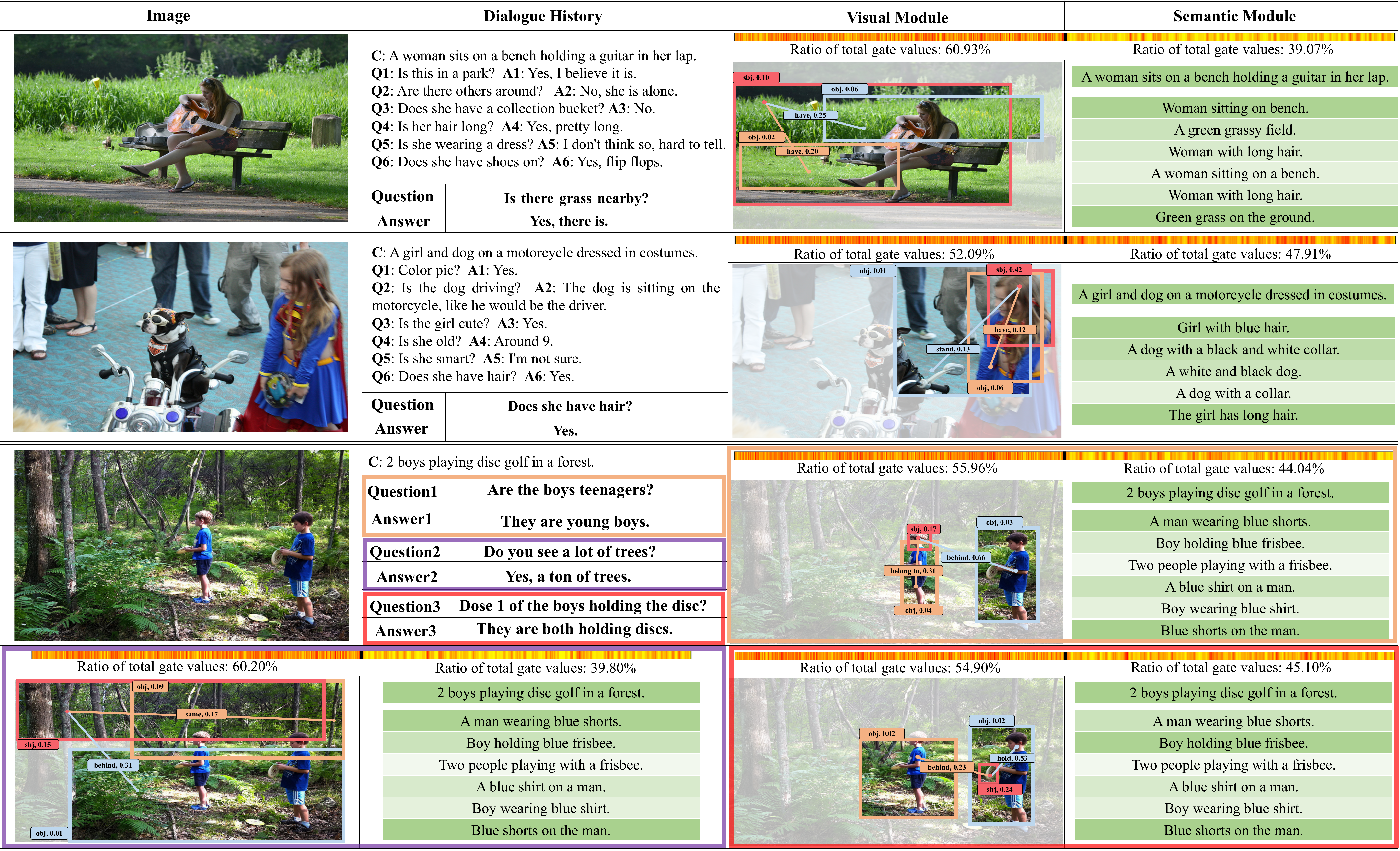} %visual2.pdf
\caption{Visualization for DualVD. %Three examples are shown. 
Visual module highlights the most relevant subject (red box) according to attention weights of each object ($\gamma_i^v$ in Eq. \ref{objectAtt}) and the objects (orange and blue boxes) with the top two attended relationships ($\beta _{ij}$ in Eq. \ref{relationAtt}). Semantic module shows the attention distribution ($\delta_i^q$ in Eq. \ref{languageAtt}) over the global caption (first row) and the local captions (rest rows), where darker green color indicates bigger attention weight. The yellow thermogram on the top
%(the red is high, the yellow is low) 
visualizes the gate values ($gate^s$ in Eq. \ref{fussAtt}) of the visual embedding (left) and the caption embedding (right) in visual-semantic fusion. The ratio of gate values for the visual module and semantic module is computed from Eq. \ref{fussAtt}.}
%defined as $G_v/G_t$ and $G_s/G_t$ (where $G_v$ is visual gate values, $G_v$ is semantic gate values and $G_t$ is total gate values in Eq. \ref{fussAtt}).}

\label{result_visual}
\end{figure*}

\iffalse
\begin{table}[t] 
\caption{Ablation study on ELMo of DualVD on VisDial v1.0 validation set (the first block) and test-standard set (the second block). }
\label{elmo_ab}
%\begin{center} 
\begin{tabular}{L{1.8cm}C{0.6cm}C{0.6cm}C{0.6cm}C{0.69cm}C{0.6cm}C{0.77cm}}%{llllllX}   
%\begin{tabular}{L{1.5cm}L{0.65cm}L{0.65cm}L{0.65cm}L{0.65cm}L{0.65cm}L{0.7cm}}
\hline                       
Model & MRR & R\textsl{@}1 & R\textsl{@}5 & R\textsl{@}10 & Mean & NDCG  \\
\hline  
w/o ELMo &63.67&49.89&80.44&89.84&4.14&56.41\\
DualVD& 64.64 & 50.74 & 82.10 & 91.00 & 3.91 & 57.30\\
\hline 
w/o ELMo  &62.84 & 49.13 & 79.93 & 89.15 & 4.39& 55.51\\
DualVD & 63.23 & 49.25 & 80.23 & 89.70 & 4.11 & 56.32\\

\hline  
\end{tabular}  
%\end{center} 
\end{table} 

\fi

\begin{table}[t] 
\caption{Ablation study of DualVD on VisDial v1.0.}
\label{ablation}
%\begin{center} 
\resizebox{.95\columnwidth}{!}{
\begin{tabular}{L{1.66cm}C{0.6cm}C{0.6cm}C{0.6cm}C{0.69cm}C{0.6cm}C{0.77cm}}%{llllllX}   
%\begin{tabular}{L{1.5cm}L{0.65cm}L{0.65cm}L{0.65cm}L{0.65cm}L{0.65cm}L{0.7cm}}
\hline                       
Model & MRR & R\textsl{@}1 & R\textsl{@}5 & R\textsl{@}10 & Mean & NDCG  \\
\hline  
ObjRep &63.84&49.83&81.27&90.29&4.07&55.48\\
RelRep &63.63 & 49.25 & 81.01 & 90.34 & 4.07& 55.12 \\
VisNoRel & 63.97 & 49.87 & 81.74 & 90.60 & 4.00& 56.73 \\ 
VisMod &64.11& 50.04&  81.78  &   90.52 &  3.99   &56.67 \\
\hline 
GlCap & 60.02 & 45.34 & 77.66 & 87.27 & 4.78 & 50.04\\
LoCap & 60.95 &  46.43 & 78.45 & 88.17 & 4.62 &51.72\\
SemMod & 61.07 & 46.69 & 78.56 & 88.09 & 4.59 & 51.10\\
\hline 
w/o ELMo &63.67&49.89&80.44&89.84&4.14&56.41\\
\textbf{DualVD} & \textbf{64.64} & \textbf{50.74} & \textbf{82.10} & \textbf{91.00} & \textbf{3.91} & \textbf{57.30}\\

\hline  
\end{tabular} 
}
%\end{center} 
\end{table}

In Table \ref{ablation}, models in the first block are designed to evaluate the influence of key components in the visual module.  \textbf{ObjRep} only considers isolated objects and ignores the relational information, which achieves worse performance compared with VisMod. \textbf{RelRep} considers the relationships by introducing relation embedding. However, empirical study indicates that enhancing visual relationships while weakening object appearance is still not sufficient for better performance. \textbf{VisNoRel} fuses the information from both object appearance and neighborhoods without relational semantics, which achieves slight improvement compared to ObjRep. On top of VisNoRel, \textbf{VisMod} moves a step further by aggregating all the neighborhood features with relational information, which achieves the best performance compared to above three models. 

Orthogonal to visual part, models in the second block evaluate the influence of key components in the semantic part. %including global caption and local captions. 
The overall performance of either \textbf{GlCap} or \textbf{LoCap} decreases by 1\% and 0.15\% respectively, compared to their integrated version \textbf{SemMod}, which adaptively selects and fuses the task-specific descriptive clues from both global-level and local-level captions. 

\textbf{DualVD} results in a great boost compared to SemMod and a relatively slight boost compared to VisMod. This unbalanced boost indicates that visual module provides comparatively richer clues than semantic module. Combining the two modules together gains an extra boost because of their complementary information. The performance of DualVD without ELMo embedding decrease slightly, which proves that the improvement of DualVD mainly comes from the contribution of the novel visual representation.
%Except for visual and semantic, as a promising future work, we believe that incorporating semantic representations from other modalities will bring further improvements. 

\subsection{Interpretability}

A critical advantage of DualVD lies in its interpretability: DualVD is capable to predict the attention weights in the visual module, semantic module and the gate values in visual-semantic fusion. It supports explicit visualization and can reveal DualVD's mode in information selection. Figure \ref{result_visual} shows three examples with variant dependence on visual and semantic modules. The third example (third and fourth rows in Figure \ref{result_visual}) shows three round of dialogues about an image. In each round of dialogue, DualVD is capable to capture the most relevant visual and semantic information regarding the current question. In the first question, the visual module highlights the face of a boy and the relationships to his body and the other boy, while the semantic module puts more attention on the captions describing the two boys, which all provide useful clues to infer the correct answer. In the second and third round of dialogues, DualVD respectively attends to the whole grass and the discs. In this example, the attended information is adaptively changed through the dialogue and this explains why the correct answer is selected.

We further show another two examples with a current question and the dialogue history (first two rows in Figure \ref{result_visual}) to reveal DualVD's mode in information selection. We observe that the amount of information derived from each module highly depends on the complexity of the question and the relevance of the content. More information will come from the semantic module when the question involves complex relationships or the semantic module explicitly contains question-relevant clues. In Figure \ref{result_visual}, \textit{ratio of total gate values} reveals the amount of information derived from each module.  In the first example, more visual information is required. Similar observation exists for the second question in the third example. Such questions referring to object appearance depend more clues from the visual module. In the second example, the current question is about the relationship between the girl and the hair. The amount of semantic information remarkably increases since there exists explicit evidence ``\emph{The girl has long hair}''. This observation holds for the third question in the third example. Since language is a higher-level encoding of the visual content after complex reasoning involved with prior knowledge, it provides more useful clues for semantic-level questions.

\section{Conclusion}

%In this paper, we develop a novel model for the visual di- alog task. The backbone of this model is a GNN, in which each node represents a dialogue entity and the edge weights represent the semantic dependencies between nodes. An EM-style inference algorithm is proposed for this GNN to estimate the latent relations between nodes and the missing values of unobserved nodes. Experiments are performed on the VisDial and VisDial-Q dataset. Results show that our method is able to find and utilize underlying dialogue struc- tures for dialogue inference in both tasks, demonstrating the generality and effectiveness of our method.

%In this work we study visual relationship detection at an un- precedented scale and propose a novel model that can gener- alize better on long tail class distributions. We find it is cru- cial to integrate subject and object features at multiple levels for good relation embeddings and further design a loss that learns to embed visual and semantic features into a shared space, where semantic correlations between categories are kept without hurting discriminative ability. We validate the effectiveness of our model on multiple datasets, both on the classification and detection task, and demonstrate the superi- ority of our approach over strong baselines and the state-of- the-art. Future work includes integrating a relationship pro- posal into our model that would enable end-to-end training
In this paper, inspired by the dual-coding theory in cognitive science, we propose a novel DualVD model for visual dialogue. DualVD mainly consists of a visual module and a semantic module, which encodes image information at appearance-level and semantic-level, respectively. Desired clues for answer inference are adaptively selected from the two modules via gate mechanism. Results from extensive experiments on benchmarks demonstrate that deriving visual information from visual-semantic representations can achieve superior performance compared to other state-of-the-art approaches. Another major advantage of DualVD is its interpretability via progressive visualization. It can give us insight of how information from different modalities is used for inferring answers. 

\section{Acknowledgement}
This work is supported by the National Key Research and Development Program (Grant No.2017YFB0803301).

%Extensive Experiments are performed on the VisDial v0.9 and VisDial v1.0. Results show that the model is able to make the dialogue effectively by dynamically utilizing the information from vision channel and language channel which fits neurological theory as well as makes agent more intelligent and humanize.

\bibliographystyle{aaai}
\bibliography{essay.bbl}

\end{document}